# A Hybrid Approach to Word Sense Disambiguation combining Supervised and Unsupervised Learning

Alok Ranjan Pal,[1, 3] Anirban Kundu,[2, 3] Abhay Singh,[1] Raj Shekhar,[1] Kunal Sinha[1]

[1]College of Engineering and Management, West Bengal, India 721171
`{chhaandasik, abhaysingh3185, rajshekharssp, kunalsameer87}@gmail.com`

[2]Kuang-Chi Institute of Advanced Technology, Shenzhen, P. R. China 518057
`anirban.kundu@kuang-chi.org`

[3]Innovation Research Lab (IRL), West Bengal, India 711103
`anik76in@gmail.com`

## Abstract

*In this paper, we are going to find meaning of words based on distinct situations. Word Sense Disambiguation is used to find meaning of words based on live contexts using supervised and unsupervised approaches. Unsupervised approaches use online dictionary for learning, and supervised approaches use manual learning sets. Hand tagged data are populated which might not be effective and sufficient for learning procedure. This limitation of information is main flaw of the supervised approach. Our proposed approach focuses to overcome the limitation using learning set which is enriched in dynamic way maintaining new data. Trivial filtering method is utilized to achieve appropriate training data. We introduce a mixed methodology having "Modified Lesk" approach and "Bag-of-Words" having enriched bags using learning methods. Our approach establishes the superiority over individual "Modified Lesk" and "Bag-of-Words" approaches based on experimentation.*

## Keywords

*Word Sense Disambiguation (WSD), Modified Lesk (ML), Bag-of-Words (BOW).*

## 1. Introduction

In human languages all over the world, there are a lot of words having different meaning depending on the contexts. Word Sense Disambiguation (WSD) [1-5] is the process for identification of probable meaning of ambiguous words based on distinct situations. The word "Bank" has several meaning, such as "place for monitory transaction", "reservoir", "turning point of a river", and so on. Such words with multiple meaning are ambiguous in nature. The process of identification to decide appropriate meaning of an ambiguous word for a particular context is known as WSD. People decide the meaning of a word based on the characteristic points of a discussion or situation using their own merits. Machines have no ability to decide such an ambiguous situation unless some protocols have been planted into the machines' memory.

In supervised learning, a learning set is considered for the system to predict the meaning of ambiguous words using a few sentences having a specific meaning of the particular ambiguous words. Specific learning set is generated as a result for each instance of different meaning. A system finds the probable meaning of an ambiguous word for the particular context based on defined learning set. In this method, learning set is created manually unable to generate fixed rules for specific system. Therefore predicted meaning of an ambiguous word in a given context can't be always detected. Supervised learning is capable to derive partial predicted result, if the learning set does not contain sufficient information for all possible senses of the ambiguous word. It shows the result, only if there is information in the predefined database [6-7].

In unsupervised learning, online dictionary is taken as learning set avoiding the inefficiency of supervised learning. "WordNet" is the most widely used online dictionary [8-14] maintaining "words and related meanings" as well as "relations among different words".

The WSD process is important for different applications such as information retrieval [15], automated classification [16] and so on. WSD plays an important role in the field of language translation by machine [17-19].

Two typical algorithms "Lesk" [20, 21] and "Bag-of-Words" [6] are coupled in this paper with some modification.

The organization of rest of the paper is as follows: Section 2 is about the related activities of our paper, based on the existing methods; Background of the paper is briefly mentioned in Section 3; Section 4 describes the proposed approach with algorithmic description; Section 5 depicts experimental results along with comparison; Section 6 represents the conclusion of the paper.

## 2. RELATED WORK

Many algorithms have been designed in WSD based on supervised and unsupervised learning. "Lesk" and "Bag-of-Words" are two well-known methods which are discussed in this section as the basis of our proposed approach.

### 2.1. Preliminaries of Lesk

Typical Lesk approach selects a short phrase from the sentence containing an ambiguous word. Then, dictionary definition (gloss) of each of the senses for ambiguous word is compared with glosses of other words in that particular phrase. An ambiguous word is being assigned with the particular sense, whose gloss has highest frequency (number of words in common) with the glosses of other words of the phrase.

Example 1: "Ram and Sita everyday go to bank for withdrawal of money."
Here, the phrase is taken depending on window size (number of consecutive words). If window size is 3, then the phrase would be "go bank withdrawal". All other words are being discarded as "stop words".
Consider the glosses of all words presented in that particular phrase are as follows:
The number of senses of "Bank" is '2' such as 'X' and 'Y' (refer Table 1).
The number of senses of "Go" is '2' such as 'A' and 'B' (refer Table 2).
The number of senses of "Withdrawal" is 2 such as 'M' and 'N' (refer Table 3).

Table 1. Probable Sense of "Bank".

| Keyword | Probable sense |
|---------|----------------|
| Bank    | X              |
|         | Y              |

Table 2. Probable Sense of "Go".

| Word | Probable sense |
|------|----------------|
| Go   | A              |
|      | B              |

Table 3.  Probable Sense of "Withdrawal".

| Word | Probable sense |
|---|---|
| Withdrawal | M |
| | N |

Consider the word "Bank" as a keyword. Number of common words is measured in between a pair of sentences.

Table 4.  Comparison Chart between pair of sentences and common number of words within particular pair.

| Pair of Sentences | Common number of Words |
|---|---|
| X and A | A' |
| X and B | B' |
| Y and A | A'' |
| Y and B | B'' |
| X and M | M' |
| X and N | N' |
| Y and M | M'' |
| Y and N | N'' |

Table 4 shows all possibilities using sentences from Table 1, Table 2, Table 3, and number of words common in each possible pair.

Finally, two senses of the keyword "Bank" have their counter readings (refer Table 4) as follows:
X counter, $X_C = A' + B' + M' + N'$.
Y counter, $Y_C = A'' + B'' + M'' + N''$.

Therefore, higher counter value would be assigned as the sense of the keyword "Bank" in particular sentence. This strategy believes that surrounding words have same senses as of the keyword

## 2.2. Preliminaries of Bag-of-Words

The Bag-of-Words approach is a model, used in Natural Language Processing (NLP), to find out the actual meaning of a word having different meaning due to different contexts. In this approach, there is a bag for each sense of a keyword (disambiguated word) and all the bags are manually populated. When the meaning of a keyword would be disambiguated, the sentence (containing the keyword) is picked up and the entire sentence would be broken into separate words. Then, each word of the sentence (except "stop words") would be compared with each word of each "sense" bags searching for the maximum frequency of words in common.

## 3. BACKGROUND

This paper adopts the basic ideas from typical Lesk algorithm and Bag-of-Words algorithm introducing some modifications.

## 3.1 Modified Lesk Approach

In this approach, gloss of keyword is only considered within specific sentence instead of selection of all words. Number of common words is being calculated between specific sentence and each dictionary based definitions of particular keyword.

- Consider, earlier mentioned sentence of "Example 1" as follows: "Ram and Sita everyday go to bank for withdrawal of money."
- The instance sentence would be "Ram Sita everyday go bank withdrawal money" after discarding the "stop words" like "to", "for", and so on.
- If "Bank" is considered as keyword and its two senses are X and Y (refer Table 1). Then, number of common words should be calculated between the instance sentence and each probable senses of "Bank" (refer Table 1).
- Number of common words found would be assigned to the counter of that sense of "Bank". Consider, X-counter has the value I' and Y-counter has the value I".
- Finally, the higher counter value would be assigned as the sense of the keyword for the particular instance sentence.
- The dictionary definition (gloss) of the keyword would be taken from "WordNet".
- This approach also believes that entire sentence represents the particular sense of the keyword.

## 3.2 Bag-of-Words Approach

A list of distinct words from the "Lesk" approach and "Bag-of-Words" approach is prepared based on successful disambiguation of the keyword.

- The proposed algorithm keeps unmatched words in a temporary database.
- The particular sense is being assigned to other unmatched words within temporary database based on the derivation of the sense of the ambiguous word using either of the algorithms.
- If typical "Lesk" and "Bag-of-Words" algorithms derive same sense of a particular ambiguous word, then the sense assigned to unmatched words is moved to the associated "sense bag" of the "Bag-of-Words" approach for participating directly in disambiguation.
- Else, the sense assigned to unmatched words is moved to an "anticipated database".
- If the occurrence of an unmatched word having a particular sense crosses the threshold value within the "anticipated database", then the words are considered for decision making. Therefore, the particular word is moved to the proper "sense bag".

Disambiguation probability would be increased based on enrichment of the bag. It means that learning method is tried to introduce within the typical concept of bags. If the bag grows infinitely, then disambiguation accuracy would be near to 100% in a typical way. The actual growth of the bag is limited depending on real-time memory management.

## 4. PROPOSED APPROACH

Our proposed approach is based on the "Modified Lesk" and "Bag-of-Words" approaches which are already defined in Section 3. Design of our approach is presented in form of flow chart and algorithms in this section. This approach is designed to achieve a disambiguated result with higher precision values.

In our approach, "stop words" like 'a', 'an', 'the', etc. are being discarded from input texts as these words meaningless to derive the "sense" of the particular sentence. Then, the text containing meaningful words (excluding the stop words) is passed through "Bag-of-Words" and "Modified Lesk" algorithms in a parallel fashion. "Bag-of-Words" algorithm is considered as "Module 1"; and, "Modified Lesk" is considered as "Module 2". These two algorithms are responsible to find the actual sense of ambiguous words in the particular context. The unmatched words in both these algorithms are being stored in a temporary database for further usage. After that, results of "Module 1" and "Module 2" have been being analysed to formulate the particular sense depending on the context of the sentence in "Module 3". If at least either of the algorithms (using "Module 1" or "Module 2") find the sense applying logical "OR" operation on the projected results, then particular sense is assigned to the unmatched words in the temporary database. Correctness of results based on the implemented algorithms is checked in "Module 4". If both algorithms derive same result obtained by applying "AND" operation on two results of "Module 1" and "Module 2", then the sense is considered as disambiguated sense. Therefore, unmatched words (kept in a temporary database) has to be moved to related sense bag as per the "Bag-of-Words" algorithm in "Module 1" to participate in disambiguation method now onwards. Otherwise, the derived senses are considered as the probable senses and unmatched words are being moved to an anticipated database in "Module 5". Figure 1 shows the modular division of our proposed approach. If the occurrence of a word in the anticipated database with a particular sense crosses specified threshold, the word is considered to be used for decision making and is moved to the related sense bag of the "Bag-of-Words" algorithm in "Module 1" to participate in disambiguation.

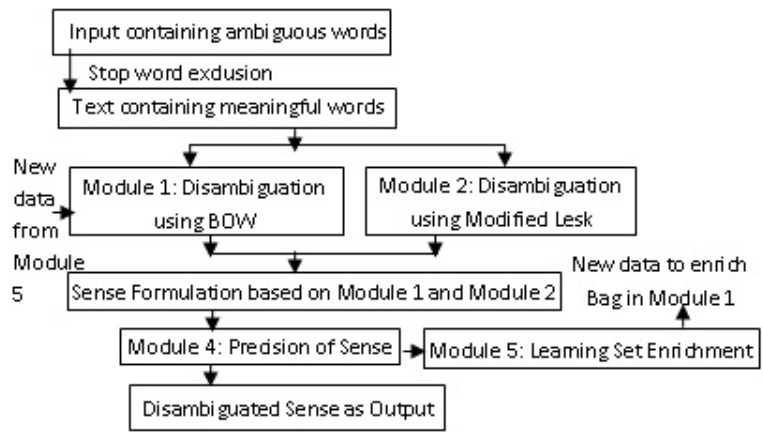

Figure 1. Modular Division of Proposed Design

Algorithm 1 is the overall procedure for the disambiguation of words (refer Figure 1). Each module performs a particular task which is mentioned in next algorithms.

**Algorithm 1: Word_Sense_Disambiguation_Process**

Input: Text containing ambiguous word
Output: Text with derived sense of ambiguous word to achieve disambiguated word
Step 1: Input text is submitted.
Step 2: All stop words like 'a', 'an', 'the', etc. are erased.
Step 3: Text with only meaningful words, are passed to Module 1 & Module 2.
Step 4: The sense of an ambiguous word is formulated in Module 3.
Step 5: Correctness of the derived sense is checked in Module 4.

Step 6: The disambiguated sense is achieved as result; and, learning set is enriched in Module 5 using new data available in Module 4.
Step 7: Stop.

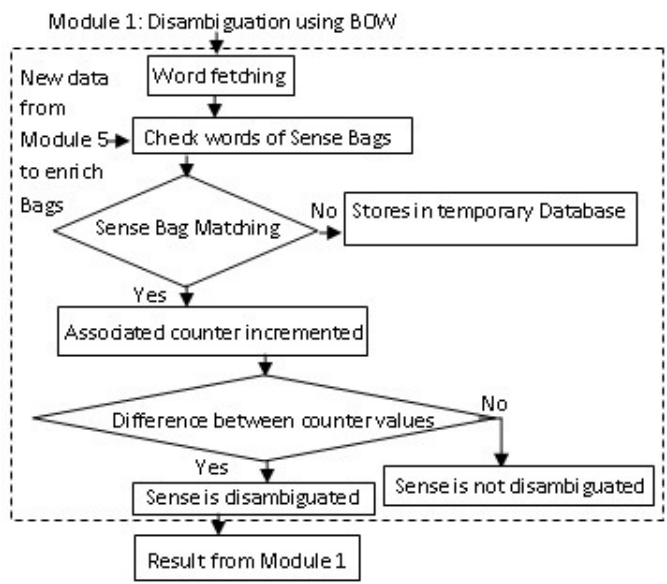

Figure 2. Flowchart of Bag-of-Words approach

Algorithm 2 is based on Module 1 and it tries to find the sense of an ambiguous word using Bag-of-Words approach (refer Figure 2).

**Algorithm 2: Find_Sense_in_Bag_of_Words**

Input: Text with only meaningful words
Output: Actual sense of ambiguous words
Step 1: Loop Start for each meaningful word of input texts.
Step 2: Each word is selected from preliminary input texts.
Step 3: If the word is matched with the word of any sense bags, then associated counter is increased.
Step 4: Else, unmatched word is stored in a temporary database.
Step 5: Loop End
Step 6: If the counter value is mismatched with all other values, then associated sense is considered as the disambiguated sense.
Step 7: Else, Bag-of-Words algorithm fails to disambiguate the sense.
Step 8: Stop.

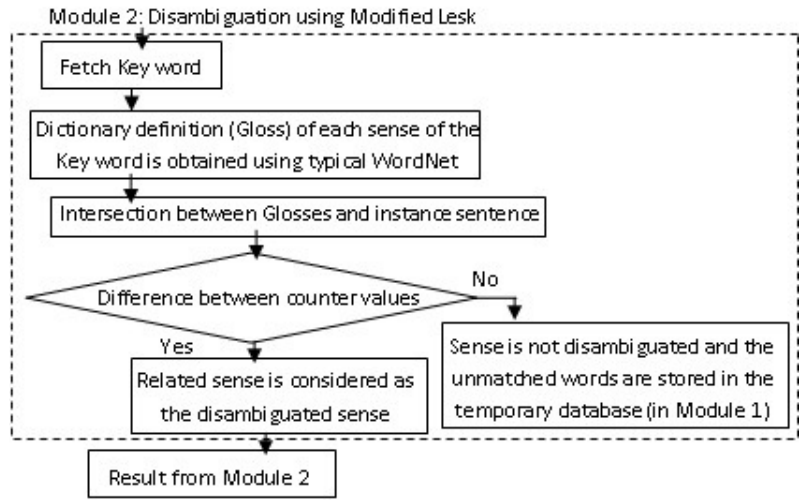

Figure 3.  Flowchart of Modified Lesk approach

Algorithm 3 is based on Module 2 and it finds the sense of an ambiguous word using Modified Lesk (refer Figure 3).

**Algorithm 3: Find_Sense_in_Modified_Lesk**

Input: Text with only meaningful words
Output: Actual sense of the ambiguous word
Step 1: Loop Start for all glosses (dictionary definitions) of the ambiguous word.
Step 2: Ambiguous word is selected.
Step 3: Gloss of ambiguous word is obtained from typical WordNet.
Step 4: Intersection is performed between the meaningful words from the input text and the glosses of the ambiguous word.
Step 5: Loop End
Step 6: If If the counter value is mismatched with all other values, then associated sense is considered as the disambiguated sense.
Step 7: Else, Modified Lesk algorithm fails to sense disambiguated word; and, unmatched words are stored in temporary database (in Module 1).
Step 8: Stop.

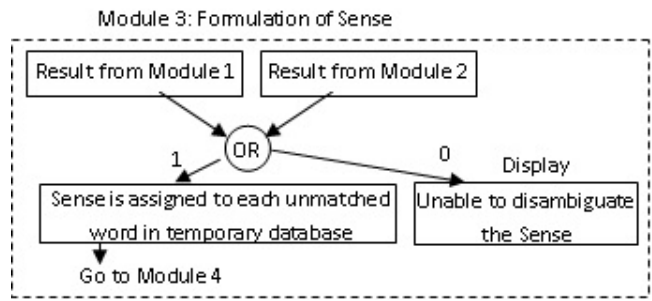

Figure 4.  Flowchart to Formulate Sense

Algorithm 4 is designed based on Module 3. It formulates actual sense of the ambiguous word using results from previous two modules (refer Figure 4). If at least one of the two approaches can derive the sense, that is considered as the disambiguated sense.

**Algorithm 4: Sense_Formulate**

Input: Results from Module 1 and Module 2
Output: Result of "OR" operation
Step 1: "OR" operation is applied on two results of Module 1 and Module 2.
Step 2: Check whether the derived sense is disambiguated by at least by Module 1 or Module 2.
Step 3: If result is '1', it means that the sense is obtained from at least one of the algorithms or both of the algorithms. Then, the particular sense is assigned to each of unmatched words within temporary database. Then, go to Module 4.
Step 4: Else, both the algorithms fail to disambiguate the sense.
Step 5: Stop.

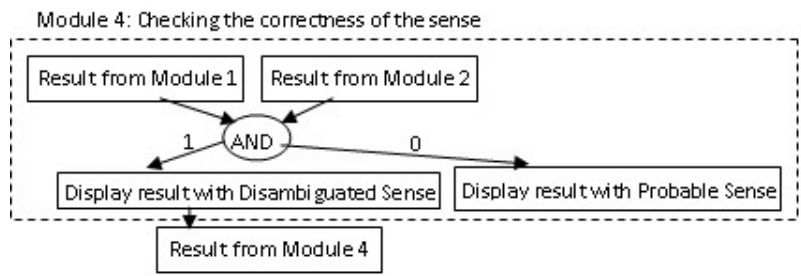

Figure 5. Flowchart for checking correctness of sense

Algorithm 5 is designed based on Module 4. It finds the correctness of disambiguated sense (refer Figure 5) using "AND" operation, derived by Module 1 and Module 2. If both approaches derive same sense, the result of "AND" operation is '1'. Otherwise, for all other cases, the result is '0'.

**Algorithm 5: Find_Sense_Precision**

Input: Results from Module 1 and Module 2
Output: Result of "AND" operation
Step 1: Collect the results of Module 1 and Module 2
Step 2: "AND" operation is applied on the results of Module 1 and Module 2
Step 3: If result is 1 (both the approaches produce same results), then derived sense is displayed as the disambiguated sense.
Step 4: Else, derived sense is displayed as a probable sense.
Step 5: Stop.

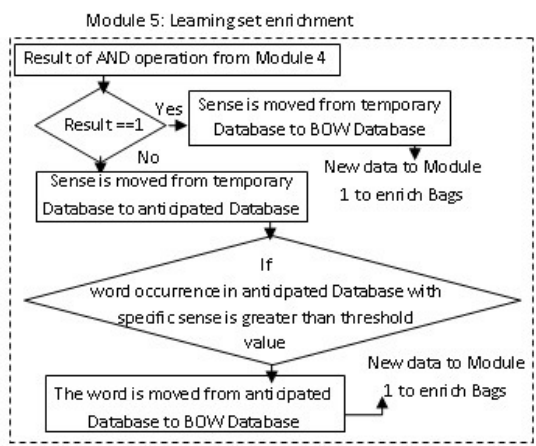

Figure 6. Flowchart for Learning Set Enrichment

Algorithm 6 is designed based on Module 5 activities. It enriches the learning set by populating with words from temporary database (refer Figure 6).

**Algorithm 6: Learning_Set_Enrichment**

Input: Sense assigned unmatched words from temporary database
Output: Enriched learning set
Step1: Result of "AND" operation from Module 4 is received as input.
Step 2: If the result is 1, then sense assigned unmatched words from temporary database are moved to specific BOW database.
Step 3: Else, sense assigned unmatched words are moved from temporary database to an anticipated database.
Step 4: If occurrence of an unmatched word in anticipated database having a particular sense crosses the threshold value, then the word is moved to the related BOW database.
Step 5: Stop.

The key feature of this algorithm is based on the auto enrichment property of the learning set. For the first time, if any word is not present in the learning set, it could not be able for participation for disambiguation. Though, its probable meaning would be stored in the database. When the number of occurrences of the particular word with a particular sense crosses specific threshold value, the word is inserted in the learning set to take part in disambiguation procedure. Therefore, the efficiency of the disambiguation process is increased by this auto increment property of the learning set.

## 5. EXPERIMENTAL RESULT

Typical word sense disambiguation based approaches examine efficiency based on three parameters such as "Precision", "Recall", and "F-measure" [20]. Precision (P) is the ratio of "matched target words based on human decision" and "number of instances responded by the system based on the particular words". Recall value (R) is the ratio of "number of target words for which the answer matches with the human decided answer" and "total number of target words in the dataset". F-Measure is evaluated as "(2*P*R / (P+R))" based on the calculation of Precision and Recall value. Different types of datasets are being considered in our experimentation to exhibit the superiority of our proposed design.

Testing has been performed on huge datasets among which a sample is considered for showing the comparison results between typical approaches and our proposed approach. In Table 5, "Plant" and "Bank" have considered as target words. Main focus is the precision value as it is the most dependable parameter in this type of disambiguation tests. Comparison among three algorithms has been depicted in Table 5.

**Sample Data for Test 1:**

This is SBI bank. He goes to bank. Ram is a good boy. Smoke is coming out of cement plant. He deposited Rs. 10,000 in SBI bank account. Are you near the bank of river? He is sitting on bank of river. We must plant flowers and trees. To maintain environment green, all must plant flowers and trees in our locality. The police made a plan with a motive to catch thieves with evidence.

Target Words: Bank, Plant.

Table 5. F-Measure Comparison in Test 1.

| Algorithms | Precision | Recall Value | F-Measure |
|---|---|---|---|
| Modified Lesk | 1.0 | 0.3 | 0.5 |
| Bag-Of-Words | 1.0 | 0.67 | 0.80 |
| Proposed Approach | 1.0 | 0.88 | 0.94 |

**Sample Data for Test 2:**

We live in an era where bank plays an important role in life. Bank provides social security. Money is an object which makes 90% human beings greedy but still people deposit money in bank without fear. Reason for above activity is trust. The bank which creates maximum trust in the hearts of people is considered to be most successful bank. Few such trustful names in India are SBI, PNB and RBI. RBI is such a big name that people can bank upon it. Here is a small story, one day a boy found a one rupee coin near the bank of the river. He wanted to keep that money safe. But he could not found any one upon whom he can bank upon. He thought to deposit the money under a tree, in the ground, near the bank of river. Moral of the story kids find earth as the safest bank. Here is another story about a beggar. A beggar deposited lot of money in her hut which was near the bank of Ganga. One day other beggars found her asset and they planned to loot that money. When the beggar came to know about the plan she shouted for help. Nobody but a bank came to rescue and they helped the 80 year old to open an account and keep her money safe.

Target Word: Bank.

Table 6. F-Measure Comparison in Test 2.

| Algorithms | Precision | Recall Value | F-Measure |
|---|---|---|---|
| Modified Lesk | 0.83 | 0.45 | 0.58 |
| Bag-Of-Words | 0.71 | 0.45 | 0.55 |
| Proposed Approach | 0.77 | 0.6 | 0.68 |

In Table 6, the result is below our expectations as initial database is small for "Bag-of-Words" approach. "Modified Lesk" (unsupervised) has shown better results than "Bag-of-Words" (supervised).

**Sample Data for Test 3:**

This is PNB bank. He goes to bank. He was in PNB bank for money transfer. He deposited Rs 10,000 in PNB bank account. Are you near the bank of river? He is sitting on bank of river. He was in PNB bank for money transfer. We must plant flowers and trees. He was in PNB bank for money transfer. This is PNB bank. This is PNB bank. This is PNB bank. He was in PNB bank for money transfer. He was in PNB bank for money transfer. He was in PNB bank for money transfer. He was in PNB bank for money transfer. This is PNB bank. This is PNB bank. This is PNB bank. This is PNB bank. This is his SBI bank.

Target Words: Bank.

Table 7. F-Measure Comparison in Test 3.

| Algorithms | Precision | Recall Value | F-Measure |
|---|---|---|---|
| Modified Lesk | 1.0 | 0.15 | 0.26 |
| Bag-Of-Words | 1.0 | 0.45 | 0.62 |
| Proposed Approach | 1.0 | 0.85 | 0.92 |

In Table 7, the text is long enough to give combined approach more chances to show its efficiency. Few lines are repeated in order to overcome the threshold value.

**Sample Data for Test 4:**

Mango plant grows in five year. Cement plant cause pollution. Mango plant can be planted in garden. Mango plant can grow into tree within five years. Building of cement plant is toughly built. Police had a plant in the terrorist gang.
Mango plant grows in five year. Cement plant cause pollution. Mango plant can be planted in garden. Mango plant can grow into tree within five years. Building of cement plant is toughly built. Police had a plant in the terrorist gang.
Mango plant grows in five year. Cement plant cause pollution. Mango plant can be planted in garden. Mango plant can grow into tree within five years. Building of cement plant is toughly built. Police had a plant in the terrorist gang.
Mango plant grows in five year. Cement plant cause pollution. Mango plant can be planted in garden. Mango plant can grow into tree within five years. Building of cement plant is toughly built. Police had a plant in the terrorist gang.
Mango plant grows in five year. Cement plant cause pollution. Mango plant can be planted in garden. Mango plant can grow into tree within five years. Building of cement plant is toughly built. Police had a plant in the terrorist gang.
Target Word: Plant.

Table 8. F-Measure Comparison in Test 4.

| Algorithms | Precision | Recall Value | F-Measure |
|---|---|---|---|
| Modified Lesk | 1.0 | 0.67 | 0.80 |
| Bag-Of-Words | 1.0 | 0.60 | 0.75 |
| Proposed Approach | 1.0 | 0.93 | 0.96 |

Table 8 contains one paragraph which is repeated 5 times. This repetition helps combined approach to enrich its bag with new words. The "Bag-of-Words" approach with a fixed size bag of data is behind the "Modified Lesk" approach. It exhibits better results in proposed approach since learning dataset is being enriched.

Table 9. Average of Test Results.

| Algorithm | Precision | Recall Value | F-Measure |
|---|---|---|---|
| Modified Lesk | 0.94 | 0.41 | 0.57 |
| Bag-of-Words | 0.87 | 0.52 | 0.65 |
| Proposed Approach | 0.92 | 0.80 | 0.86 |

Table 9 shows average values of all the tests performed. Efficiency of an algorithm based on fixed size learning set is improved in this paper enriching datasets. "Bag-of-Words" and "Modified Lesk" approaches individually exhibit the "F-Measure" as 0.65 and 0.57 respectively; whereas proposed approach shows "F-Measure" as 0.86 since learning set is dynamically enriched with new context sensitive definitions of particular words after each execution

# 6. CONCLUSIONS

In this paper, our approach has established better performance in enhanced WSD technique depending on specific learning sets. The disambiguation accuracy is improved based on the enrichment of datasets having populated by new data. We have achieved better precision value, recall value, and F-Measure through extensive experimentation.

## ACKNOWLEDGEMENT

Flow cytometry data analysis system based on GPUs platform (No. JC201005280651A).

## AUTHORS

Alok Ranjan Pal has been working as an a Assistant Professor in Computer Science and Engineering Department of College of Engineering and Management, Kolaghat since 2006. He has completed his Bachelor's and Master's degree under WBUT. Now, he is working on Natural Language Processing.

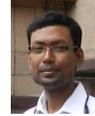

Anirban Kundu is working as Post Doctorate Research Fellow in Kuang-Chi Institute of Advanced Technology, Nanshan, Shenzhen, Guangdong, P.R.China.He worked as Head of the Department in Information Technology department, Netaji Subhash Engineering College, Garia, Kolkata, West Bengal, India.

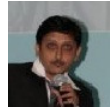

Abhay Singh is currently working at Accenture India Pvt. Ltd. as a Senior Programmer. He completed his Bachelor's degree in Information Technology from College of Engineering and Management, Kolaghat, year 2006-2010.

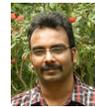

Raj Shekhar is currently working at Infosys India Ltd. as a System Engineer. He completed his Bachelor's degree in Information Technology from College of Engineering and Management, Kolaghat , year 2006-2010.

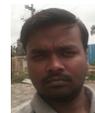

Kunal Sinha is currently working at Tata Consultancy Services Ltd. as Software Engineer. He completed his Bachelor's degree in Information Technology from College of Engineering and Management, Kolaghat, year 2006-2010.

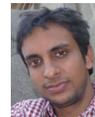